\documentclass[runningheads]{llncs}
\usepackage[T1]{fontenc}
\usepackage{graphicx,verbatim,enumitem,amsmath,amsfonts,tcolorbox,multirow,svg,booktabs,ulem,subcaption,hyperref,cleveref,graphicx}

\begin{document}
\renewcommand{\thefootnote}{}
\title{BioD2C: A Dual-level Semantic Consistency Constraint Framework for Biomedical VQA}
%

\author{
Zhengyang Ji\inst{1,2} 
\and 
Shang Gao\inst{2} 
\and 
Li Liu\inst{2}
\and
Yifan Jia\inst{1,2}
\and
Yutao Yue\inst{2,3}\textsuperscript{$\dagger$}
}

\authorrunning{Z. Ji et al.} 
\titlerunning{BioD2C}
\institute{
Shandong University, Qingdao, China
\and
The Hong Kong University of Science and Technology (Guangzhou), \\
Guangzhou, China \\
\email{yutaoyue@hkust-gz.edu.cn}
\and
Institute of Deep Perception Technology, JITRI, Wuxi, China\\
}

\maketitle              
\footnotetext{$\dagger$ Corresponding author.}
\begin{abstract}
Biomedical visual question answering (VQA) has been widely studied and has demonstrated significant application value and potential in fields such as assistive medical diagnosis. Despite their success, current biomedical VQA models perform multimodal information interaction only at the model level within large language models (LLMs), leading to suboptimal multimodal semantic alignment when dealing with complex tasks. To address this issue, we propose \textbf{BioD2C}: a novel \textbf{D}ual-level Semantic \textbf{C}onsistency \textbf{C}onstraint Framework for \textbf{Bio}medical VQA, which achieves dual-level semantic interaction alignment at both the model and feature levels, enabling the model to adaptively learn visual features based on the question. Specifically, we firstly integrate textual features into visual features via an image-text fusion mechanism as feature-level semantic interaction, obtaining visual features conditioned on the given text; and then introduce a text-queue-based cross-modal soft semantic loss function to further align the image semantics with the question semantics. Specifically, in this work, we establish a new dataset, BioVGQ, to address inherent biases in prior datasets by filtering manually-altered images and aligning question-answer pairs with multimodal context, and train our model on this dataset. Extensive experimental results demonstrate that BioD2C achieves state-of-the-art (SOTA) performance across multiple downstream datasets, showcasing its robustness, generalizability, and potential to advance biomedical VQA research. The source code of this work and the BioVGQ dataset can be accessed through \href{https://github.com/jzy-123/BioD2C}{code} and \href{https://huggingface.co/datasets/jzyang/BioVGQ}{dataset}.

\keywords{Biomedical VQA \and Dual Interaction \and Semantic Alignment}

\end{abstract}

\section{Introduction}

\begin{figure}[t]
    \centering
    \includegraphics[width=\textwidth]{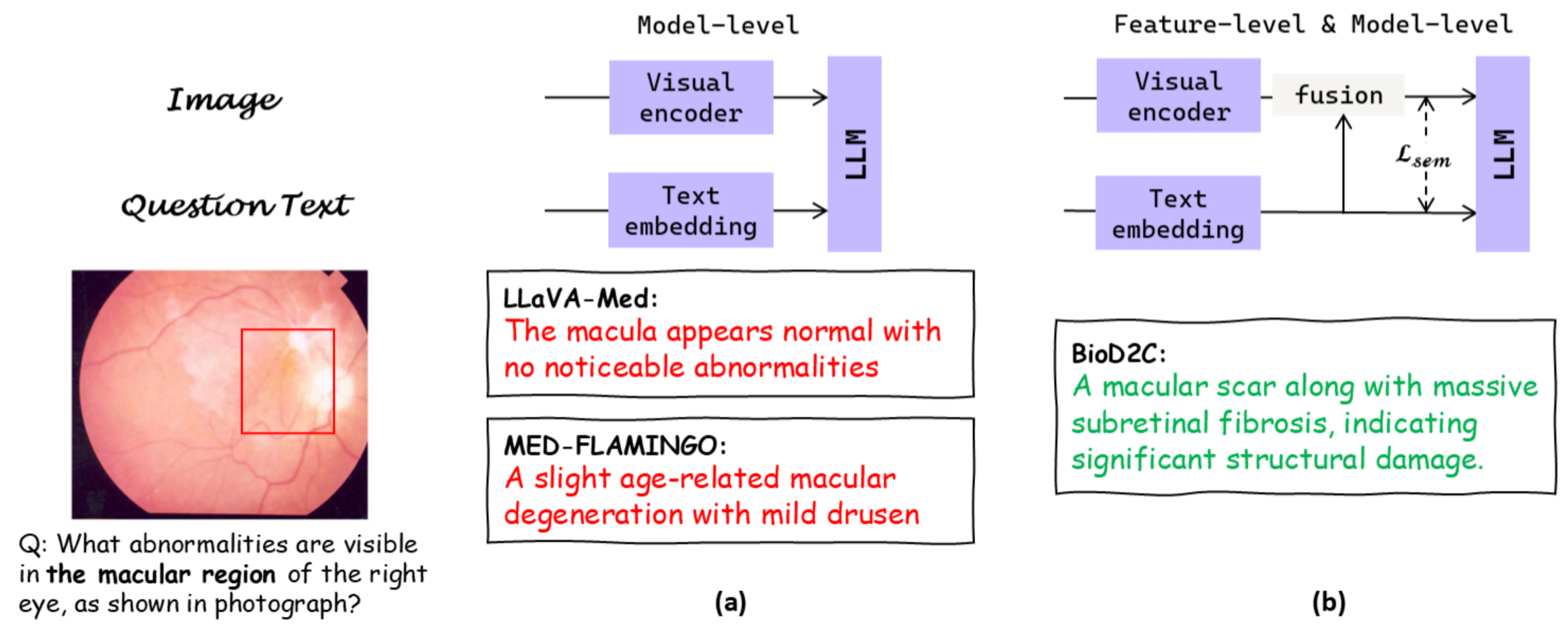}
    \caption{(a) and (b) illustrate the performance of the model-level interaction framework and BioD2C under image-related questions, respectively. Red text represents incorrect answers, while green text represents correct answers.}
    \label{intro}
\end{figure}
Biomedical VQA aims to design and develop systems capable of understanding biomedical images and generating relevant answers based on given textual instructions.

In real clinical scenarios, question texts often refer to specific elements within an image. Therefore, achieving optimal semantic alignment between the image and the text instruction, i.e., the model should focus on the image regions corresponding to the textual query, becomes the key to the success of biomedical question answering models. However, existing biomedical models \cite{zhang2023pmcvqa,li2023llavamed,moor2023flamingo,wu2023radfm} extract visual and textual features independently using separate visual encoders and text embedding layers, relying solely on LLMs for model-level multimodal semantic interaction, and lacking semantic alignment at the feature level. 

\begin{figure}[t]
\includegraphics[width=\textwidth]{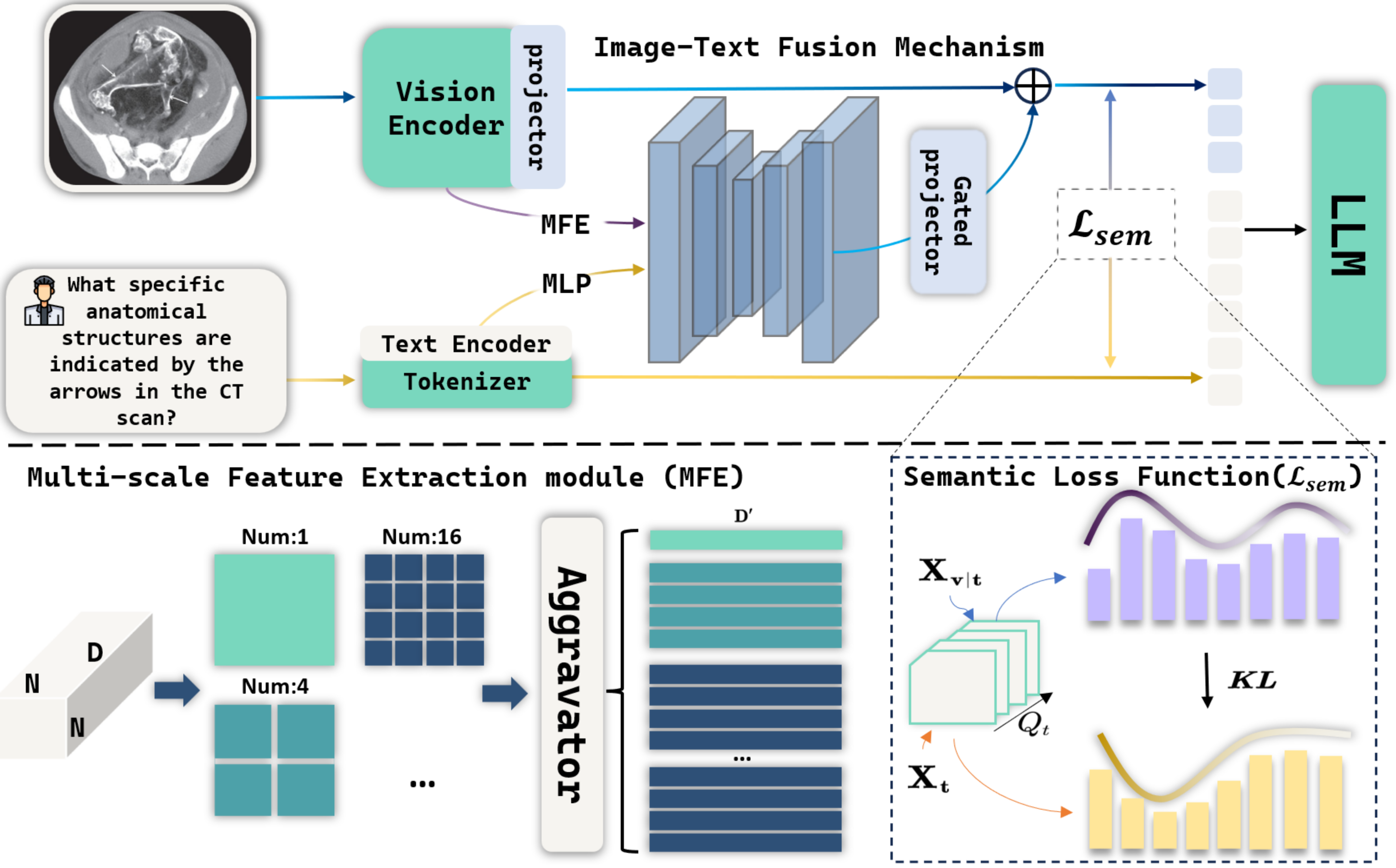}
\caption{
\textbf{BioD2C Architecture.} \textit{Feature-level Interaction:} Medical images and text questions are encoded into features $X_v$ and $X_t$.  A multi-scale enhanced $X_v$ is fused with $X_t$ via a Transformer decoder, generating $X_{vt}$, which is then combined with $X_v$ through a gating mechanism to produce text-conditioned features $X_{v|t}$. \textit{Semantic Loss:} A text-queue loss guides $X_{v|t}$ to align with $X_t$.
} 
\label{pipeline}
\end{figure}

To address these challenges, we propose BioD2C, a novel dual-level semantic consistency constraint framework for biomedical VQA, as illustrated in Fig. \ref{pipeline}. Compared to existing VQA models, BioD2C employs a novel image-text fusion mechanism for feature-level multimodal semantic interaction after extracting image and text features, obtaining image features conditioned on the given text. Furthermore, we introduce a text queue mechanism to project image and text semantics from high-dimensional vector spaces into corresponding probability distributions. By minimizing the divergence between these distributions, we achieve quantized alignment of cross-modal semantic representations. Fig. \ref{intro} illustrates this behavior. When faced with complex questions, both baseline models rely solely on semantic interaction within LLMs, resulting in biased answers, while BioD2C benefits from semantic alignment at the feature level and produces the correct answer.

Due to the scarcity of real biomedical data, existing biomedical vision-language datasets such as PMC-OA \cite{lin2023pmcoa} and PMC-VQA \cite{zhang2023pmcvqa} rely on biomedical papers publicly available from the PubMedCentral (PMC)’s OpenAccess subset \cite{roberts2001pubmed}, some of which contain images that differ significantly from real-world biomedical images. Additionally, when generating question-answer pairs, existing biomedical visual question-answering datasets often rely solely on image captions or captions supplemented with visual information, which may cause a misalignment between the generated question-answer pairs and the information contained in the images. As a result, visual question-answering models trained on these datasets may have an inherent limitation in understanding the images.

To this end, we establish a new dataset, BioVGQ, to address the issues present in existing datasets. It is based on the existing PMC-based dataset and integrates multiple public datasets, filtering out images that have undergone significant manual manipulation. When generating question-answer pairs, both the images and their corresponding captions are utilized to ensure a strong correlation between the question-answer pairs and the images.

Overall, the main contributions of our work are as follows: i) We propose BioD2C, a dual-level framework that enforces semantic consistency both through model-level interactions within LLMs and specifically through feature-level image-text fusion mechanisms, while further optimizing visual-textual alignment via a cross-modal semantic loss function. ii) BioVGQ, a biomedical VQA dataset with cleaner images that incorporate contextual information, has been established, and our model is trained on this dataset. iii) Extensive comparative and ablation experiments demonstrate the superiority of BioD2C over current SOTA biomedical VQA models in terms of performance and the effectiveness of each component.


\section{The BioVGQ Dataset}
Most of the image data in BioVGQ comes from PMC-VQA. To remove images in the original dataset that significantly differ from real medical images, we manually annotated 3000 images as either ``clean'' or ``polluted''. Using these labeled data, we trained an image classifier to automate the classification process, ultimately obtaining 77K clean biomedical images. Specifically, we added an MLP classification head to the pre-trained image encoder of PMC-CLIP \cite{lin2023pmcoa} to serve as the image classifier.

For generating biomedical question-answer pairs, we used the ChatGPT-4o API \cite{hurst2024gpt}, providing both images and their corresponding captions to ensure that the generated pairs accurately reflect the image content without deviation. Further generation details and prompts are as follows:
\begin{tcolorbox}[colback=gray!20, colframe=gray!50, boxrule=0.5mm, rounded corners]
You are an AI assistant specialized in biomedical topics. Generate 2-3 clinically meaningful open-ended question-and-answer pairs based on the provided medical image and caption.\\
Requirements:
    - Each question must be a single, clear sentence; each answer should directly address it.
    - Cover overall understanding and specific details, without copying the caption.
    - Answers must require examining the image, not just medical background knowledge.
    - Ensure clinical relevance, professionalism, and conciseness.
    - Format: \texttt{[{"question": "xxx", "answer": "xxx"}, ...]}
    
caption: \texttt{\{caption\}}, image-url: \texttt{\{image-url\}}
\end{tcolorbox}

To enrich the dataset, we incorporated various modalities of biomedical images, closed-ended questions, and short dialogues, integrating the training splits of SLAKE \cite{liu2021slake}, Path-VQA \cite{he2020pathvqa}, and RAD-VQA \cite{vqarad}. As a result, BioVGQ comprises 81K medical images and 188K question-answer pairs. 

\section{Method}
\subsection{Feature Level Semantic Interaction Mechanism}\label{model}

In this section, we will introduce the technical details of feature-level semantic interaction through the image-text fusion mechanism.

\textbf{Text Encoding}. Before fusing textual features with image features, we preprocess the text in two steps. First, the tokenized text sequence $\mathcal{T}$ is encoded into a meaningful representation $X_{t}^{'}$, defined as $X_{t}^{'}=\varepsilon \left( \mathcal{T} \right) $, where $\varepsilon$ represents the encoding function. Here, we directly use the text embedding layer of the LLM as the encoding function $\varepsilon$.  Next, we use a MLP to map $X_{t}^{'}$ into the image feature space, resulting in the final text representation $X_{t}$. The overall text encoding process is expressed as: $X_t=MLP\left( \varepsilon \left( \mathcal{T} \right) \right)$.

\textbf{Image Processing}. The visual features output by the vision encoder, $X_v\in \mathbb{R}^{N\times N\times D}$, contain only a single level of granularity, where $N$ represents the number of image patches and $D$ denotes the feature dimension. To extract multi-granularity image features, we introduce a multi-scale feature extraction module (MFE) that utilizes a divide-and-aggregate strategy, as illustrated in Fig. \ref{pipeline}. Specifically, MFE consists of $S$ different scales. At each scale $s\in \left\{ 1,2,...,S \right\}$, the feature map is divided into $4^{s-1}$ blocks, resulting in a total of $M=\sum_{s=1}^S{4^{s-1}}$ blocks across all scales. Each 3D feature block undergoes global pooling to produce a 1D feature. For a block $f_{s,t}$ at scale $s$, where $t\in \left\{ 1,2,...,4^{s-1} \right\}$ indicates the block index, the global pooling is defined as $f_{s,t}^{'}=maxpool\left( f_{s,t} \right) +avgpool\left( f_{s,t} \right)$, where $maxpool$ and $avgpool$  represent global max pooling and global average pooling, respectively. Finally, the pooled features $f_{s,t}^{'}$ from all scales are concatenated to form the multi-scale image feature $X_{v}^{'}\in \mathbb{R}^{M\times D}$. In the implementation, the number of scales is set to $S=6$.

\textbf{Image-Text Fusion}. A 12-layer Transformer decoder achieves cross-modal fusion by treating text encoding $X_t$ as query and multi-scale image features $X_{v}^{'}$ as key and value, producing the text-contextualized image representation $X_{vt}$:
\begin{equation}
X_{vt} = Fusion(X_t, X_v', X_v').
\end{equation}
 To compensate for the potential loss of original image features during modality fusion and achieve complementary information across modalities, we introduce a learnable gating mechanism \cite{aberdam2023gate1,alayrac2022gate2,gate3,graves2012gate4}. This mechanism combines the original image features $X_v$ with the fused features $X_{vt}$ by processing them through a projection layer and an additional projection layer, resulting in the conditioned image features $X_{v\left| t \right.}$. The gating mechanism ensures a gradual fusion of modality features, avoiding significant feature alteration and overall performance degradation \cite{ganz2024questionaware}. It is implemented by multiplying the output of the additional projection layer with $tanh\left( \beta \right)$, where $\beta$ is a learnable parameter initialized to a small positive value. We initialize $\beta$ to 0.2 to balance initial feature bias and fusion effectiveness. Mathematically, the fusion module is implemented as:
\begin{equation}
    X_{v\left| t \right.}=Proj\left( X_v \right) +Proj_g\left( X_{vt} \right) \cdot tanh\left( \beta \right),
\end{equation}
then $X_{v\left| t \right.}$ is input into the LLM together with the text token sequence $\mathcal{T}$.

\subsection{Text-queue-based Cross-modal Semantic Loss function}
Through the above procedure, we obtain the visual features conditioned on the text, but lack an optimization objective to guide the model toward optimal multimodal semantic alignment at the feature level. Inspired by ALBEF \cite{li2021albef} and MoCo \cite{he2020moco}, we propose a text-queue-based cross-modal semantic loss function, which applies a soft constraint to align visual semantics with text semantics. The core idea is to map the semantics from the high-dimensional vector space to a probability distribution through similarity computation. Specifically, we extract $k$ text samples semantically related to the image either from its corresponding caption or by retrieving from an existing knowledge base, where we set $k$ to 30. These texts undergo the text encoding process in Section \ref{model} to obtain the text queue $\mathcal{Q}_t=\left\{  t_i \right\} _{i=1}^{k}$, where $t_i$ denotes the $i_{th}$ text sample. By calculating the cosine similarity between $X_{v\left| t \right.}$, $X_t$, and the elements in $\mathcal{Q}_t$, we derive the semantic distributions of fused image and text features, denoted as $p\left( v \right)$ and $p\left( t \right)$, respectively. The semantic distribution $p\left( v \right)$ is computed as:
\begin{equation}
    p\left( v \right) =\left\{ \frac{\exp \left( \left< \left. X_{v\left| t \right.}, t_i  \right> /\tau \right. \right)}{\sum_{j=1}^k{\exp \left( \left< \left. X_{v\left| t \right.}, t_j  \right> /\tau \right. \right)}} \right\} _{i=1}^{k},
\end{equation}
where $\left< \cdot ,\left. \cdot \right> \right. $ represents the calculation of cosine similarity, and $\tau$ is the temperature coefficient. Similarly, $p\left( t \right)$ can be computed. Using $p\left( v \right)$ and $p\left( t \right)$, we minimize the Kullback-Leibler (KL) divergence between these two distributions to align the semantics of image and text features, expressed as: 
\begin{equation}
   \mathcal{L}_{sem}=D_{KL}\left( p\left( v \right) ||p\left( t \right) \right),  
\end{equation}
where $\mathcal{L}_{sem}$ represents the semantic loss between images and text, and $D_{KL}$ denotes the KL divergence.
During training, the semantic loss $\mathcal{L}_{sem}$ is combined with the commonly used sequence negative log-likelihood loss $\mathcal{L}_{nll}$ \cite{sutskever2014nll} to jointly optimize the model for the best performance. The final loss function used for training the model is defined as:
\begin{equation}
    \mathcal{L}_{total}=\lambda \cdot \mathcal{L}_{sem}+\mathcal{L}_{nll},
    \label{loss}
\end{equation}
where $\lambda$ is a hyperparameter that controls the weight of the semantic loss.

\section{Experiments}
\subsection{Implementation Details}
In this work, we employ a two-stage training strategy to train our model, enabling it to adapt to biomedical VQA tasks. \textbf{Stage 1:} Projectors are independently trained to align visual features with language embeddings using 467k image-caption pairs from the LLaVA-Med dataset \cite{li2023llavamed}. During this stage, $\lambda=0$ in Eq. \ref{loss}, disabling semantic loss due to limited textual diversity. \textbf{Stage 2:} The LORA adapters are fine-tuned on the BioVGQ dataset to improve BioD2C’s multimodal understanding, with $\lambda=1$ in Eq. \ref{loss}, fully incorporating semantic loss to optimize performance.

We train our models using the AdamW \cite{loshchilov2017adamw} optimizer. To accelerate training, we employ the Deepspeed strategy along with Automatic Mixed Precision (AMP) \cite{feng2021AMP} and gradient checkpointing. Set the learning rates for the first and second training stages to $\left\{ 5e-5,\ 2e-5 \right\} $, and train for 1 epoch and 5 epochs, respectively. For more details on hyperparameter settings, please refer to the BioD2C GitHub page. All models are implemented in PyTorch and trained on four NVIDIA 4090 GPUs with 24 GB of memory each. In terms of model construction, PMC-CLIP and PMC-LLaMA \cite{wu2024pmcllama} are selected as the visual encoder and LLM, respectively.

\subsection{Datasets and Metrics}

The BioVGQ dataset is split into training, validation, and testing sets in an 8:1:1 ratio for model training and evaluation. To validate the effectiveness of BioD2C, we evaluate it on SLAKE, Path-VQA, and RAD-VQA datasets. 
For ablation studies and visualization analyses, we primarily use the BioVGQ test set to examine the effectiveness of different modules.

We use closed-ended question accuracy(ACC), open-ended question ACC, BLEU-1 score \cite{papineni2002bleu}, and ROUGE-1 score \cite{lin2004rouge} to comprehensively evaluate the model's performance on downstream datasets. Additionally, as BioVGQ primarily contains long-text answers, ChatGPT4 \cite{achiam2023gpt4} is employed to evaluate the reasonableness, accuracy, and similarity of the model's responses compared to the ground truth. A comprehensive score ranging from 0 to 10 is provided, with higher scores indicating better model performance.

\begin{table}[t]
    \centering
    \resizebox{\textwidth}{!}{
    \begin{tabular}{cccccccc}
        \toprule
        \textbf{Dataset} & \textbf{Metric} & \textbf{BioMedGPT} & \textbf{LLaVA-Med-1.5}& \textbf{MedVInT-TD}& \textbf{RadFM} & \textbf{BiMediX2-8B}& \textbf{BioD2C} \\
        \midrule
        \multirow{3}{*}{SLAKE} & closed ACC $\uparrow$&  0.248&  0.536&  0.498&  0.752&  \textbf{0.831}&  \underline{0.763}\\
        & opened ACC $\uparrow$&  0.259&  0.334&  0.338&  0.725&  \underline{0.729}&  \textbf{0.742}\\
        & BLEU-1 $\uparrow$&  0.175&  0.002&  0.213&  0.746&  \textbf{0.778}&  \underline{0.766}\\
        & ROUGE-1 $\uparrow$&  0.26&  0.413&  0.351&  0.695&  \underline{0.786}&  \textbf{0.810}\\
        \midrule
        \multirow{3}{*}{RAD-VQA} & closed ACC $\uparrow$
&  0.545&  0.547&  0.475&  0.577&  \underline{0.725}&  \textbf{0.734}\\
        & opened ACC $\uparrow$
&  0.14&  0.276&  0.195&  \textbf{0.335}&  0.305&  \underline{0.310}\\
        & BLEU-1 $\uparrow$
&  0.033&  0.021&  0.125&  0.475&  \textbf{0.552}&  \underline{0.520}\\
        & ROUGE-1 $\uparrow$&  0.372&  0.342&  0.235&  0.438&  \underline{0.565}&  \textbf{0.588}\\
        \midrule
        \multirow{3}{*}{Path-VQA} & closed ACC $\uparrow$
&  0.512&  0.621&  0.454&  0.505&  \underline{0.872}&  \textbf{0.918}\\
        & opened ACC $\uparrow$
&  0.053&  0.036&  0.022&  0.005&  \underline{0.282}&  \textbf{0.291}\\
        & BLEU-1 $\uparrow$
&  0.021&  0.011&  0.013&  0.257&  \underline{0.587}&  \textbf{0.620}\\
        & ROUGE-1 $\uparrow$&  0.287&  0.116&  0.034&  0.221&  \underline{0.593}&  \textbf{0.628}\\
        \midrule
        \multicolumn{2}{c}{\textbf{Average}}&  0.242&  0.271&  0.246&  0.453&  \underline{0.616}&  \textbf{0.641}\\
        \bottomrule
    \end{tabular}
    }
    
    \caption{Comparison of performance with SOTA models on different benchmarks. The best performance is highlighted in \textbf{bold}, while the second-best is \uline{underlined}.}
    \label{tab:sota}
\end{table}
\subsection{Comparison with SOTAs}
We compare the proposed BioD2C with SOTA models in the biomedical visual question answering domain, including BioMedGPT \cite{zhang2024biomedgpt}, LLaVA-Med-1.5, MedVInT \cite{zhang2023pmcvqa}, RadFM \cite{wu2023radfm}, and BiMediX2-8B \cite{mullappilly2024bimedix2}. The results are shown in Table \ref{tab:sota}. BioD2C outperforms current SOTA models on most metrics, achieving the highest average score of 0.641, which is a 4.06\% improvement over the second-best model, BiMediX2-8B.

\subsection{Ablation Study}
In this section, we conduct ablation studies to analyze the impact of different model configurations on its performance in biomedical question answering tasks. Specifically, we investigate the following three scenarios, i) $w/o\ \mathcal{L}_{sem}$: not using semantic loss, ii) $w/o\ fm$: directly using the visual encoder's output as the image vector for LLMs multimodal input without the fusion mechanism, and iii) $w/o\ BioVGQ$: using the PMC-VQA dataset instead of BioVGQ in the second training stage. The results of the ablation studies are shown in Table \ref{tab:abl}.
\begin{table}[t]
    \centering
    \footnotesize
    \begin{tabular}{lcccc}
        \toprule
        & BLEU-1 & ROUGE-1 & GPT score/10 & Average \\
        \midrule
        $\text{BioD2C\ }_{w/o\ \mathcal{L}_{sem}}$ & 0.408& 0.443& 0.623& 0.491\\
        $\text{BioD2C\ }_{w/o\ fm}$ & 0.371& 0.428& 0.591& 0.463\\
        $\text{BioD2C}_{\ w/o\ BioVGQ}$ & 0.324& 0.332& 0.483& 0.380\\
        $\text{BioD2C\ }$ & 0.427& 0.494& 0.649& 0.523\\
        \bottomrule
    \end{tabular}
    \caption{The performance of BioD2C and its variants on the BioVGQ test set.}
    \label{tab:abl}
\end{table}

\begin{figure}[t]
    \centering
    \begin{minipage}{0.63\textwidth}
        \centering
        \includegraphics[width=\linewidth]{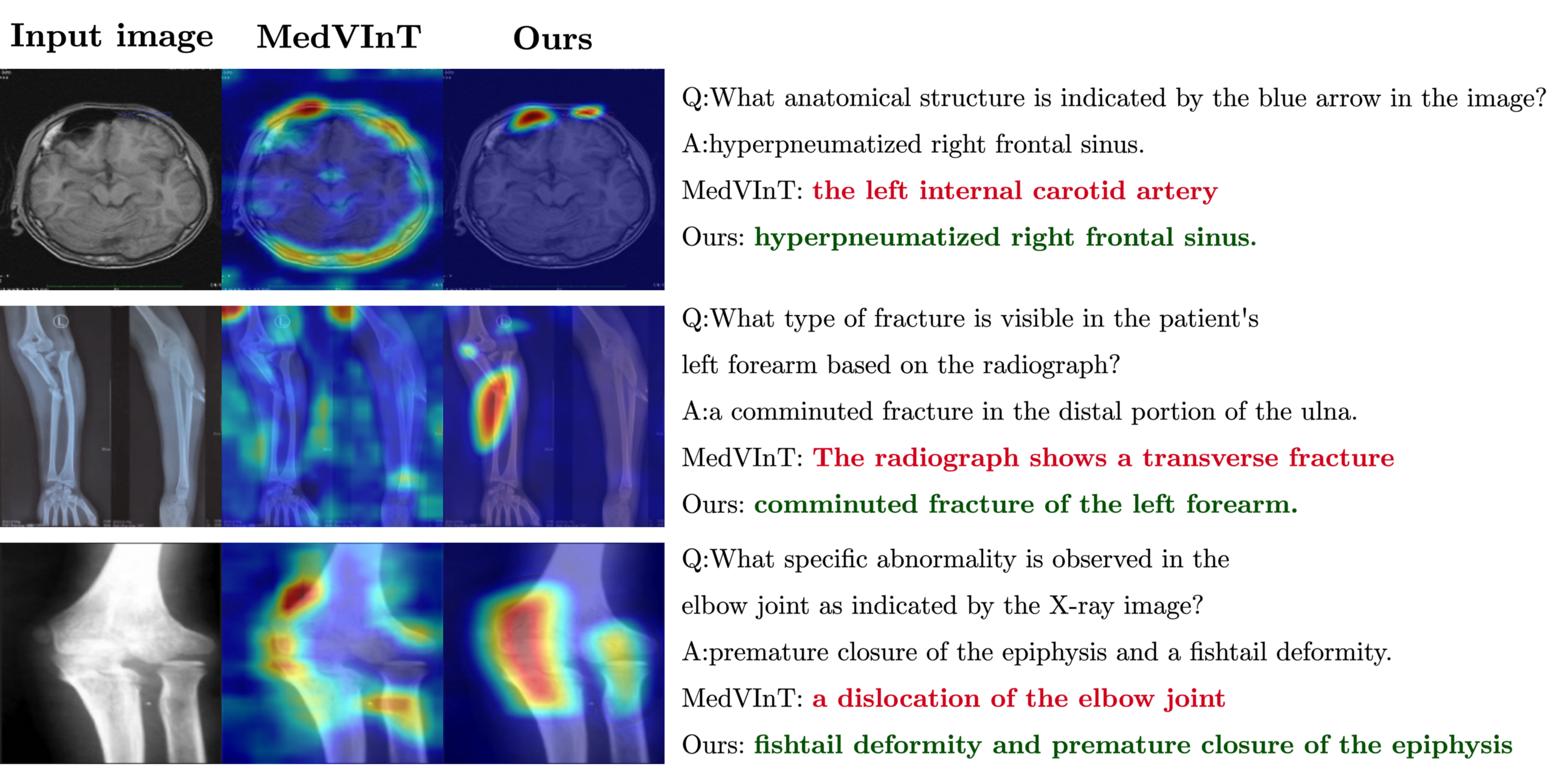}
    \end{minipage} \hfill
    \begin{minipage}{0.35\textwidth}
        \centering
        \includegraphics[width=\linewidth]{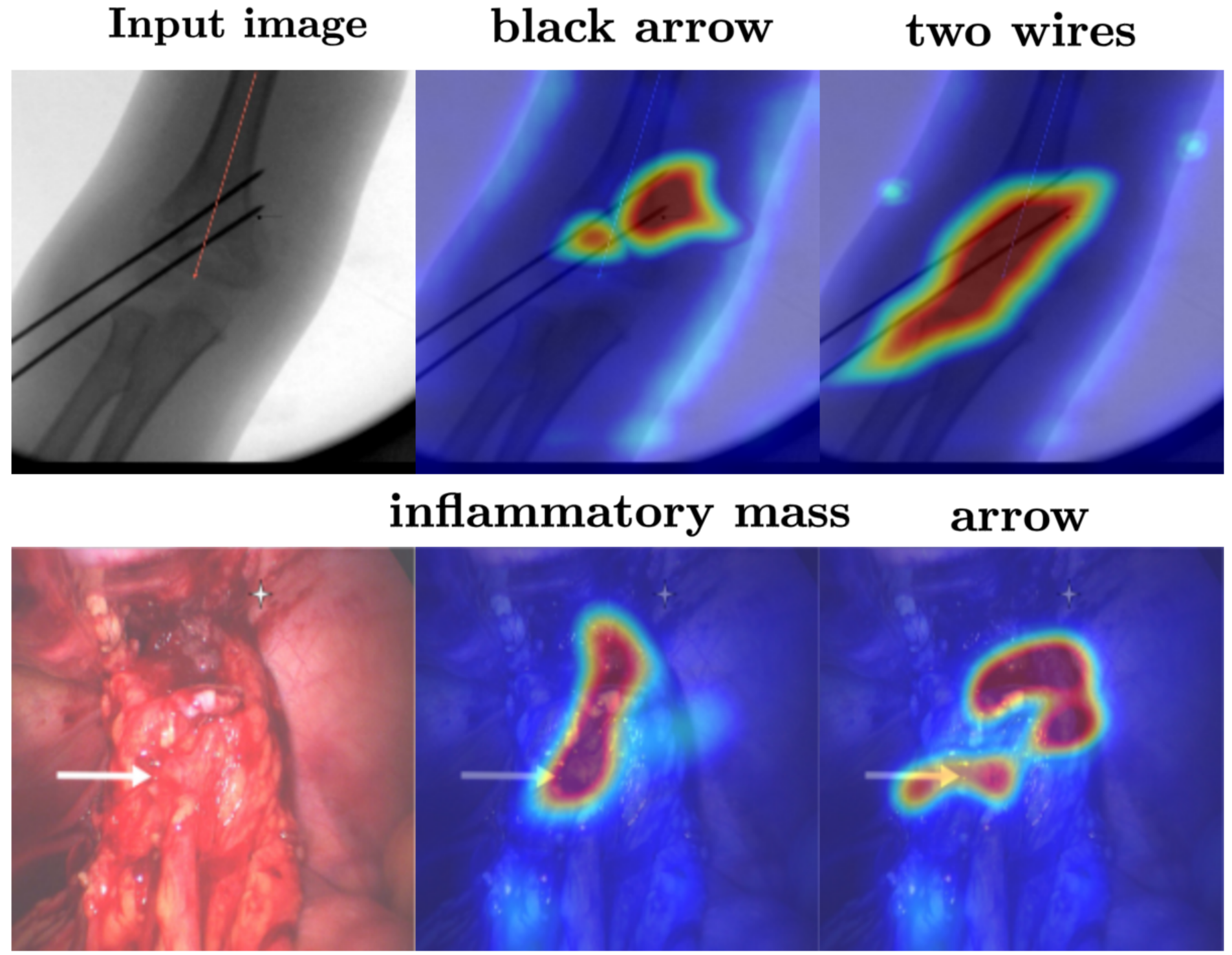}
    \end{minipage}
    \caption{Visualization of the attention map of the input image.}
    \label{vis}
\end{figure}
\subsection{Visualization Analysis}

In this section, by visualizing the attention maps, we show how the model focuses on specific regions of the image based on textual instructions. The left figure in Fig. \ref{vis} compares the performance of BioD2C and the baseline model. While both understand abstract image concepts, BioD2C accurately focuses on image regions for correct answers, while the baseline model’s answers deviate due to multi-modal alignment issues. The right figure illustrates the dynamic nature of the model’s attention: as the text prompt changes, the attention shifts to different image regions. For example, when the question mentions ``\textbf{black arrow}'', the model's attention focuses on the area near the black arrow. When the prompt changes to ``\textbf{two wires}'', the model shifts its attention to the region where the wires are located.

\section{Conclusion}
In this work, we propose BioD2C, a dual-level semantic interaction biomedical VQA framework, which achieves dynamic alignment of visual features to textual features at the feature level through an image-text fusion mechanism. A cross-modal semantic loss function is employed to further optimize multimodal semantic alignment at the feature level. The framework is trained on BioVGQ, a curated dataset consisting of 81K images and 188K question-answer pairs. Extensive experiments demonstrate that, compared to baselines that independently extract visual and textual features, BioD2C can dynamically focus on specific regions of the image based on the text, achieving SOTA performance. BioD2C shows strong potential for clinical decision support, with future work targeting multi-modal integration and broader medical applications.

\section{Acknowledgments}
This work is supported by the Guangzhou-HKUST(GZ) Joint Funding Program (Grant No. 2023A03J0008) and the Education Bureau of Guangzhou Municipality.
%
%
%
\bibliographystyle{splncs04}
\bibliography{mybibliography}
%
\end{document}